\documentclass[10pt,twocolumn,letterpaper]{article}

\usepackage{cvpr}              

\usepackage[dvipsnames]{xcolor}

\definecolor{cvprblue}{rgb}{0.21,0.49,0.74}
\usepackage[pagebackref,breaklinks,colorlinks,citecolor=cvprblue]{hyperref}
\usepackage{gensymb}
\usepackage{multirow}
\usepackage {comment}
\usepackage{stfloats}

\newcommand{\RN}[1]{
  \hspace{-0.2em}{\textbf{\romannumeral#1})}\hspace{-0.4em}
}

\newcommand\customparagraph[1]{
    \vspace{0.2em}\noindent\textbf{#1}
}

\def\paperID{*****} 
\def\confName{CVPR}
\def\confYear{2024}

\title{F$^3$Loc: Fusion and Filtering for Floorplan Localization}

\author{
    Changan Chen$^{1,*}$\quad Rui Wang$^{2}$\quad Christoph Vogel$^{2}$\quad Marc Pollefeys$^{1,2}$\vspace{3mm} \\
    $^{1}$ETH Zürich\quad $^{2}$Microsoft Mixed Reality \& AI Lab Zürich \\
    \normalsize{$^*$Work done during his internship at Microsoft Mixed Reality \& AI Lab Zürich}\\
    \normalsize{Project Page : \url{https://felix-ch.github.io/f3loc-page/}}
}

\begin{document}
\twocolumn[{%
\renewcommand\twocolumn[1][]{#1}%
\maketitle
\vspace{-10mm}
\begin{center}
    \centering
    \captionsetup{type=figure}
    \includegraphics[width=0.8325\textwidth]{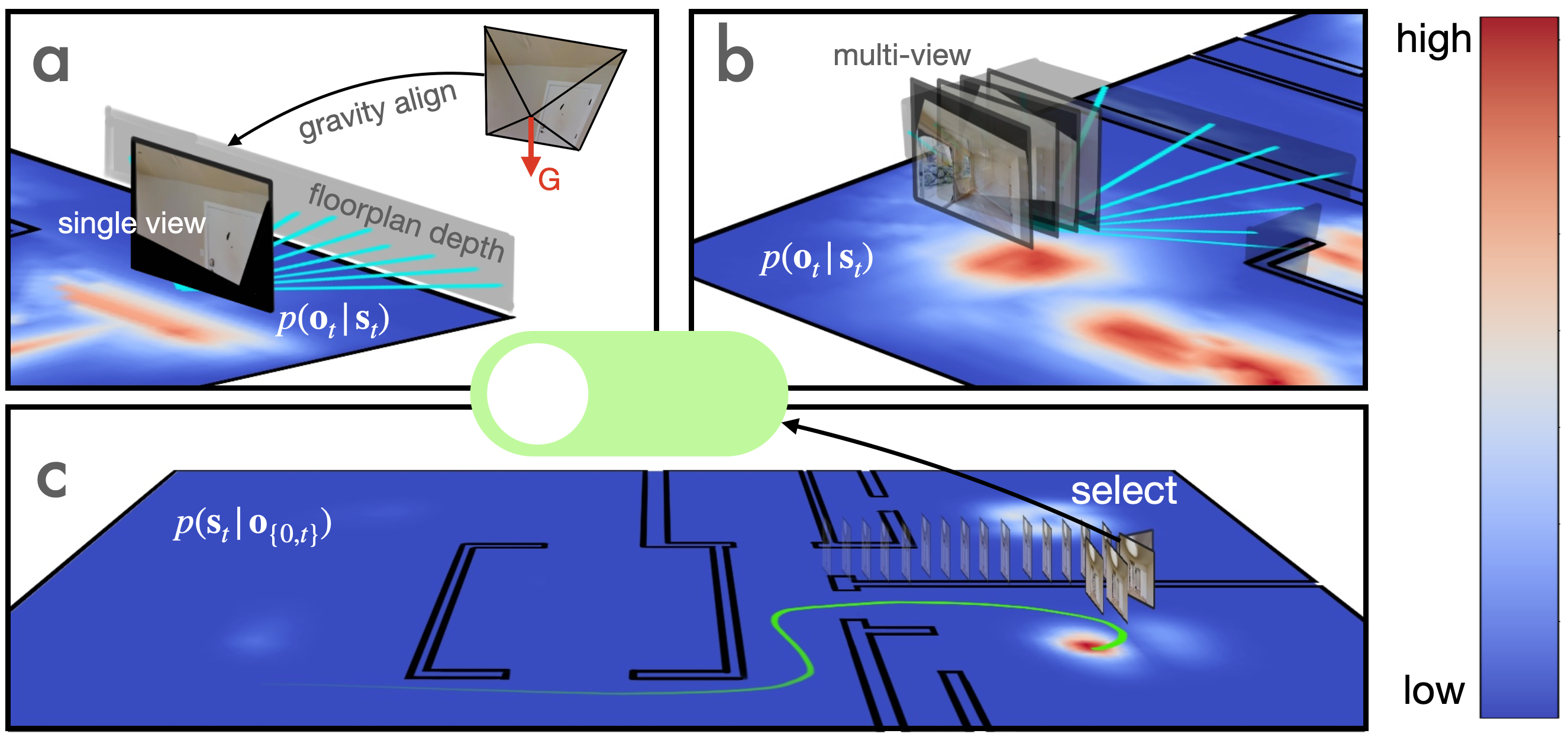}
    \captionof{figure}{\textbf{Floorplan localization.} 
    We propose a novel probabilistic model for localization within a floorplan consisting of a data-driven observation (a,b) and a temporal filtering module (c). 
    Evidence is estimated as a 1D-range image from a single (a) and a few consecutive RGB images (b).
    A learned soft selection module combines the output from the complementary cues. 
    The observation likelihood is integrated over time by an efficient SE2 histogram filter to deliver the pose posterior.
    Our system achieves rapid and accurate sequential localization, outperforming the state-of-the-art in recall and localization speed, while operating on consumer hardware.
    }
\end{center}%
}]

\begin{abstract}
In this paper we propose an efficient data-driven solution to self-localization within a floorplan.
Floorplan data is readily available, long-term persistent and inherently robust to changes in the visual appearance.
Our method does not require retraining per map and location or demand a large database of images of the area of interest.
We propose a novel probabilistic model consisting of an observation and a novel temporal filtering module.
Operating internally with an efficient ray-based representation, 
the observation module consists of a single and a multiview module to predict horizontal depth from images and fuses their results to benefit from advantages offered by either methodology.
Our method operates on conventional consumer hardware and overcomes a common limitation of competing methods~\cite{pfnet, lalaloc, lalaloc++, laser} that often demand upright images.
Our full system meets real-time requirements, while outperforming the state-of-the-art~\cite{pfnet, laser} by a significant margin.
\end{abstract}

\section{Introduction}
Camera localization is an essential research topic in computer vision.  It is key to many AR/VR applications for head-mounted or handheld mobile devices and is of great practical interest to the robotics community. Most existing works localize the camera using a pre-collected database \cite{cityscale}\cite{netvlad}\cite{relocnet} or within a pre-built 3D model~\cite{2d3d1,2d3d2,2d3d3,2d3d4,hloc}. However, these representations of the environment are costly in terms of storage and maintenance.  In contrast, indoor environments including most commercial real estate such as warehouses, offices and apartments already possess a floorplan. The floorplan is a generic representation for indoor environments that is easily accessible, lightweight, and preserves long-term scene structure independent of a changing visual appearance, such as the furnishing of the scene. It encodes rich enough information that humans can localize in an unvisited scene with its help. Therefore, we propose to localize the camera with respect to a given floorplan. This cannot only be used for indoor AR/VR applications such as floorplan navigation but also empowers robot autonomy in indoor exploration, navigation as well as search and rescue~\cite{rescue}. Our framework can be used complementary to indoor SLAM, where it can provide an initial guess for camera relocalization and significantly simplify detecting and verifying loop closures.

Due to its simple and compact form, floorplans contain many repetitive structures such as corners and walls. This causes ambiguity in the localization~\cite{lalaloc, lalaloc++, laser}, which can be eliminated to a certain extent by using image sequences~\cite{orienter, lamar}. However, incorporating the single frame localization into a sequential filtering framework~\cite{kalmanFilter,particleFilter} is challenging. The single frame localization needs to be accurate and its efficiency is crucial to ensure a high frequency of the filter with a large amount of samples~\cite{pfnet, laser}. To tackle these challenges, we propose a data-driven multi-view geometry based localization framework, that is both fast and accurate. Furthermore, we integrate this framework into a novel and highly efficient histogram filter that outputs a probability over poses and, thus, allows for multiple hypotheses in ambiguous environments but integrates evidence over time to resolve such ambiguity.

Most of the existing work assumes an upright camera pose~\cite{pfnet, lalaloc, lalaloc++, laser}, while some methods~\cite{lalaloc, lalaloc++} explicitly only consider panorama images.
In contrast, our method is designed to work with low-cost sensors, \eg, those readily available in all modern phones.
Our framework takes only a single perspective image per time-step but operates at a high speed to allow for the frequent integration of new data.
To cope with poses with non-zero roll-pitch angle, we utilize the data of an inertial measurement unit and propose a novel data augmentation method 
to overcome the limitation of previous methods~\cite{pfnet, lalaloc, lalaloc++, laser}.

In this paper we propose the following contributions. \hfill\\
\RN{1} We base our model on a novel 1D ray representation that reflects the 2D floorplan representation. 
\RN{2} We extract scene geometry from single and multi-view cues. A novel selection network fuses them in dependence of the current relative poses to take advantage of either methodology.
\RN{3} A data augmentation technique using virtual roll-pitch overcomes the limitations of current state-of-the-art methods and allows to cope with non-zero roll-pitch angles in practical use cases.
\RN{4} To eliminate ambiguity and boost localization, the predictions are filtered over time by a novel and efficient histogram filter formulated as grouped convolution from ego-motion.
\RN{5} Our full system outperforms the state-of-the-art methods in both accuracy and efficiency on existing benchmarks and a real world experiment further illustrates its potential for practical applications.
\RN{6} We collect a large indoor dataset, composed of floorplans and both short and long sequential observations in 119 Gibson~\cite{igibson} indoor environments. The dataset will be released publicly.

\section{Related Work}

\customparagraph{Visual Localization} is one of the oldest problems in computer vision and is addressed by using various methodologies. 
Image retrieval based methods \cite{cityscale}\cite{netvlad}\cite{relocnet} find the most similar image in a database and estimate the query image pose using the pose of the retrieved one. 
Methods based on a pre-built 3D SfM model of the environment~\cite{2d3d1,2d3d2,2d3d3,2d3d4,hloc} establish 2D-3D correspondences between a query image and the 3D structure by matching local descriptors and compute the image pose using minimal solvers and RANSAC. 

Recent data driven models deviate from these classical pipelines.
Scene coordinate regression \cite{dsac}\cite{forest}\cite{forest1} learns to regress the 3D coordinates of the pixels in the query image.
Pose regression methods \cite{posenet}\cite{lstmpose}\cite{branchpose} use a neural network to directly regress a 6D camera pose from the input image.
These methods rely on a pre-built 3D model that requires large storage and are scene-specific, which renders them unable to handle unvisited environments.

Instead of using a 3D model to recover the full 6D camera pose, some works tackle localization with an overhead image, such as a map \cite{orienter, geolocation}, a satellite patch\cite{vigor, crossview} or a floorplan\cite{lalaloc, lalaloc++, laser} to estimate the SE2 camera pose or R2 camera location. 
These methods can localize in unvisited scenes as long as some form of map is provided.

\customparagraph{Floorplan localization} is often associated with Lidar localization \cite{lidar1, lidar2, lidar3, lidar4, sedar}.
However, the use of Lidar inhibits the usability on common mobile devices.
Similar geometric cues can be obtained from other sources, such as point cloud reconstruction from a depth camera~\cite{wifi} or Visual Odometry (VO)~\cite{mimick}.
\cite{edge} extract room edges and compare them against the floorplan layout. 
To reconstruct 3D geometry, these works usually assume the knowledge of room or camera height~\cite{mimick, edge}.
Recently, learning-based methods use only RGB images to localize in a floorplan.
LaLaLoc~\cite{lalaloc} estimates the position of a panorama image in a given floorplan. 
Assuming known camera and ceiling height, panoramic depth images are rendered at sampled positions within the floorplan.
Localization is achieved by comparing map and image features that are embedded into the same feature space during training.
%
%
LaLaLoc++~\cite{lalaloc++} eliminates the assumption of known camera and ceiling height by directly embedding the entire floorplan into the feature space. 
Laser~\cite{laser} represents the floorplan as a set of points and gathers features, embedded by Pointnet~\cite{pointnet}, of the visible points for each pose in the floorplan. Images are embedded into a circular feature lying in the same space as the pose features. Similar to LaLaLoc and LASER, our framework actively compares rendered pose features and query image features to localize. 

PF-net~\cite{pfnet} tackles visual floorplan localization within a differentiable particle filtering framework. 
Its observation model is a learned similarity between the image and the corresponding front-facing map patch. 
The entire system is end-to-end trainable. 
However, their observation model does not appear as strong as those in~\cite{lalaloc, lalaloc++, laser}. 
\begin{figure*}[t!]
    \includegraphics[width=0.96\textwidth]{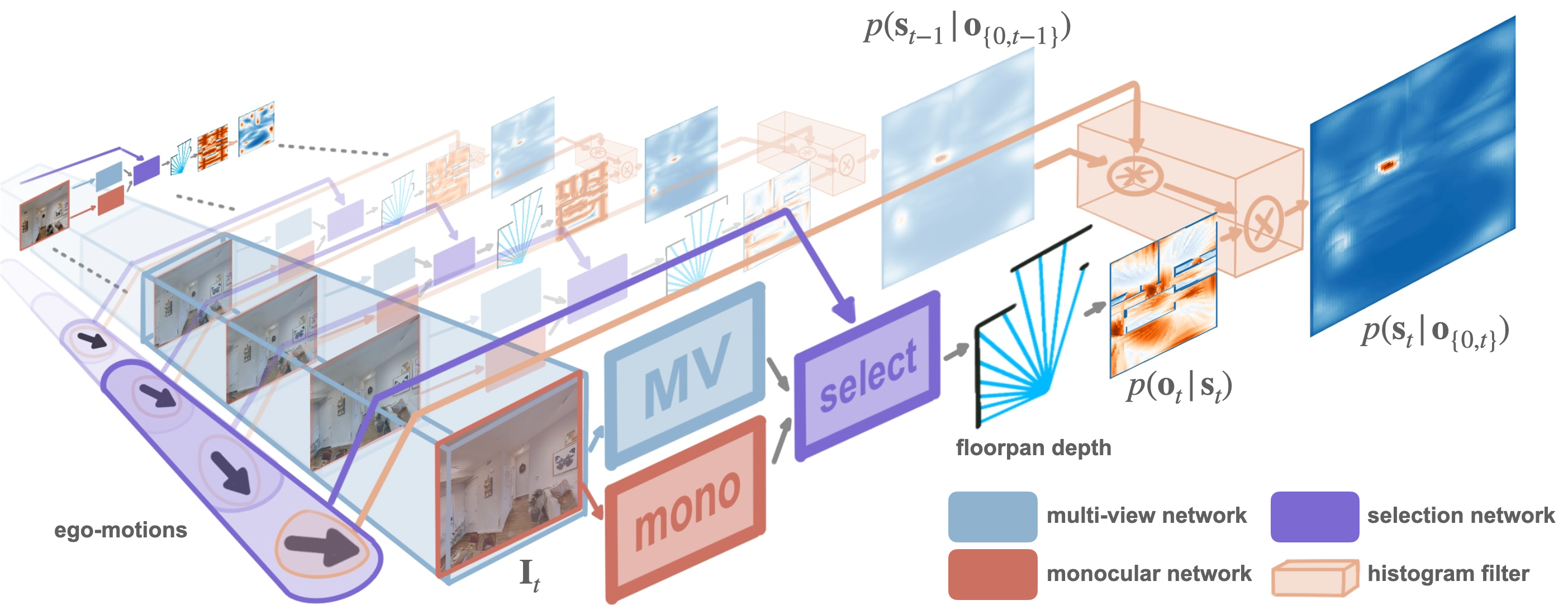}
    \caption{\textbf{Pipeline overview.} Our pipeline adopts a monocular~(\cref{sec:mono}) and a multi-view network~(\cref{sec:mv}) to predict floorplan depth. 
    A selection network~(\cref{sec:select}) consolidates both predictions based on the relative poses. 
    The resulting floorplan depth is used in our observation model and integrated over time by our novel SE(2) histogram filter~(\cref{sec:histogram}) to perform sequential floorplan localization.}
    \label{fig:overview}
\end{figure*}

\cite{lalaloc, lalaloc++, laser, pfnet} all assume that the images are captured with an upright camera pose. 
This is a strong requirement for devices such as head-mounted or hand-held devices and appears impractical for some VR/AR use cases. 
Particularly, LaLaLoc~\cite{lalaloc} and LaLaLoc++~\cite{lalaloc++} only work with panorama images, restricting their deployment on most mobile devices. 
In contrast, we propose a data augmentation scheme to cope with non-upright camera poses, improving the practicability of the method. 
Furthermore, our method utilizes 1D-range images as internal representation, instead of unorganized point cloud data, 2D-depth or RGB images.

LASER~\cite{laser} and SeDAR~\cite{sedar} use semantic information such as windows and doors as additional source of information.
Because such data is not always present in any floorplan, we consider only occupancy information in this work.

\customparagraph{Sequential localization}, \ie integrating predictions over time can increase the robustness against the observation model, eliminate scene ambiguities and boost the performance of localization~\cite{lamar, orienter}.
A common framework for fusing sequential observations is the Bayesian filter~\cite{mcl, sedar, pfnet, edge, wifi, mimick}, which maintains the posterior distribution of the current pose in an online fashion.
Implementations differ in the representation of the posterior, which can be Gaussian belief
(Kalman Filter~\cite{kalmanFilter}), a histogram (Histogram Filter~\cite{hist}) or weighted particles (Particle Filter~\cite{particleFilter}). 
As mentioned, PF-net~\cite{pfnet} introduces the particle filter specifically for floormap localization.
Here we argue that a histogram filter allows for more scaleable and effective filtering.
\cite{hist} consider the measurement update as elementwise multiplication, and transition as convolution.
However, the presented 1D and 2D cases are not practical for our localization tasks that require at least SE2 pose estimation.
To this end, we propose to consider the SE2 motion update as grouped convolution with transition filters derived from known ego-motion that can be implemented efficiently.

\customparagraph{Depth Estimation} provides strong geometric information for localization.
Recent advances in deep learning have enabled dense depth prediction from a single 
image~\cite{eigen1, eigen2, leftright, deepv2d, midas, vitmono}.
However, monocular depth estimation can suffer from scale ambiguity.
In contrast, given sufficient baseline between views, Multi-view stereo (MVS)~\cite{planesweep} does not suffer from this problem.
Current data-driven MVS methods~\cite{mvsnet, rmvsnet, pmvsnet, attnmvs, adaptmvs, unimvs} use neural networks to extract features and learn to filter a cost volume.
Instead of estimating pixel-wise depth we predict the floorplan depth of each column of the most recent gravity-aligned image, 
which can be compared directly against the floorplan.
Moreover, we benefit from the advantages of either technique by learning to fuse their predictions.

\section{Method}

\subsection{Problem Definition and Overview}
We solve the problem of localizing RGB images with respect to a floorplan. Given a temporal sequence of $k+1$ RGB images $\mathcal{I}=\left\{\mathbf{I}_\tau|\tau\in\left\{t-k,\cdots, t\right\}\right\}$ with known relative poses, camera intrinsics, and gravity directions, we aim to find the current $SE(2)$ camera pose $\mathbf{s}_t$ within a given 2D floorplan, where $\mathbf{s}_t=[s_{x,t}, s_{y,t}, s_{\phi,t}]$ represents the camera $x, y$ coordinate in the floorplan and its orientation. 
We assume the floorplan to encode necessary geometric occupancy information such as doors and walls but no semantic classes. An example is illustrated in~\cref{fig:mono} b.

We first estimate the floorplan depth (\ie, the depth to the floorplan occupancy) from the current (\cref{sec:mono}) and a few recent frames with known relative poses (\cref{sec:mv}). An MLP fuses the two estimations based on the relative poses and their respective mean depth prediction (\cref{sec:select}). 
We interpolate equiangular rays from the floorplan depth before using them to localize within the floorplan. 
A histogram filter efficiently fuses the current with integrated past belief, through grouped convolution (\cref{sec:histogram}) to deliver the final localization. 
The pipeline is illustrated in \cref{fig:overview}.

\subsection{Single Image Localization}
\label{sec:mono}
\begin{figure}[t!]
    \centering \begin{subfigure}{0.66\linewidth}
        \includegraphics[width=\linewidth]{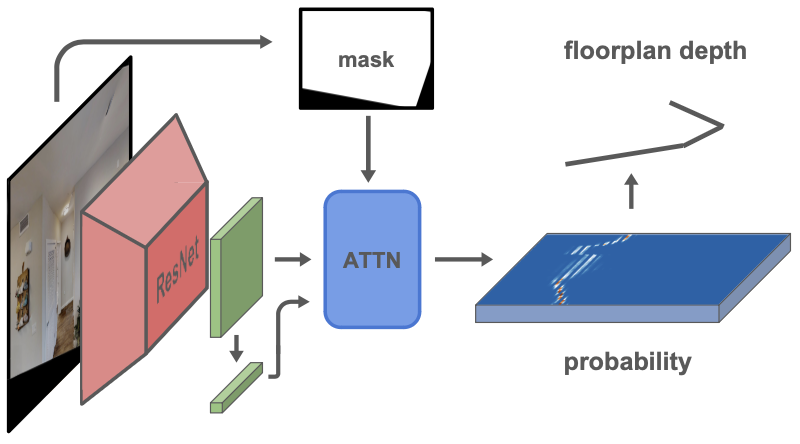}
    \caption{Monocular floorplan depth prediction.}
    \end{subfigure}
    \begin{subfigure}{0.3\linewidth}
    \includegraphics[width=\linewidth]{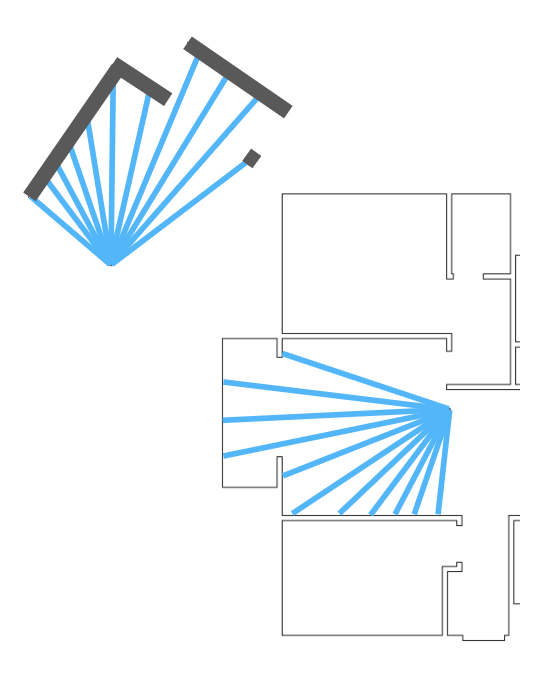}
    \caption{Localize with rays.}
    \end{subfigure}
    \caption{\textbf{Predicting and localizing with a single image.} (a) A gravity aligned image is fed into the ResNet~\cite{resnet} and Attention~\cite{attention} based feature network. Invisible pixels are masked out in the attention. The network outputs a probability distribution over depth hypotheses and its expectation is used as predicted floorplan depth. (b) Equiangular rays are interpolated from the predicted floorplan depth. We localize by finding the pose in the floorplan that has the most similar rays as the prediction.}
    \label{fig:mono}
\end{figure}

We first align the image with the gravity direction and use a ResNet\cite{resnet} and Attention~\cite{attention} based network to learn a probability distribution of the floorplan depth over a range of depth hypotheses. Pixels that become unobservable by the gravity alignment are masked out in the attention. 
The expectation is used as the floorplan depth prediction as illustrated in \cref{fig:mono}~a. 
Finally, we construct an equiangular ray scan from the predicted floorplan depth to localize in the floorplan, compare \cref{fig:mono}~b. 
The more compact representation renders the descriptor independent of the acquisition device and allows for the offline construction of the map pose features, i.e., via a circular equiangular ray scan.

\subsection{Multiview Stereo Estimation}
\label{sec:mv}
Inspired by multiview stereo, we adopt a variant of the MVS network~\cite{mvsnet, deeptam} to estimate the floorplan depth from multiple frames with known relative poses. 
We first extract features of the image columns using a ResNet~\cite{resnet} and Attention~\cite{attention} based network, and a gravity alignment mask is used in the attention.
With multiple depth hypotheses, the column features from different views are gathered via plane sweeping into the reference frame. 
This procedure is commonly used in dense multiview depth prediction~\cite{mvsnet} with the exception that we reduce our depth prediction and features vertically instead of predicting depth and extracting features for every pixel. 
Details can be found in the supplementary material.

The cross-view feature variance forms a cost distribution over the depth hypothesis. We incorporate the observability of the features at different depth hypotheses to compute meaningful variance.
Unlike traditional multiview stereo methods \cite{mvsnet, rmvsnet, pmvsnet, attnmvs, adaptmvs, unimvs} that construct 3D (without the channel dimension) cost volumes, we yield 2D cost distribution. As a consequence, the learned cost filter is 2D convolution instead of 3D. A soft-argmin computes the final floorplan depth from the filtered cost distribution as
\begin{equation}
    d = \mathbf{d}_\textrm{hyp}^\top \text{softmax}(-\mathbf{c}),
\end{equation}
where $\mathbf{d}_\textrm{hyp}\in\mathbb{R}^D$ is the vector containing the $D$ depth hypotheses, $\mathbf{c}\in\mathbb{R}^D$ is the cost at each hypothesis, and $\text{softmax}(-\mathbf{c})$ is the probability of each hypothesis.

\begin{figure}[t]
    \centering
    \includegraphics[width=0.8\linewidth]{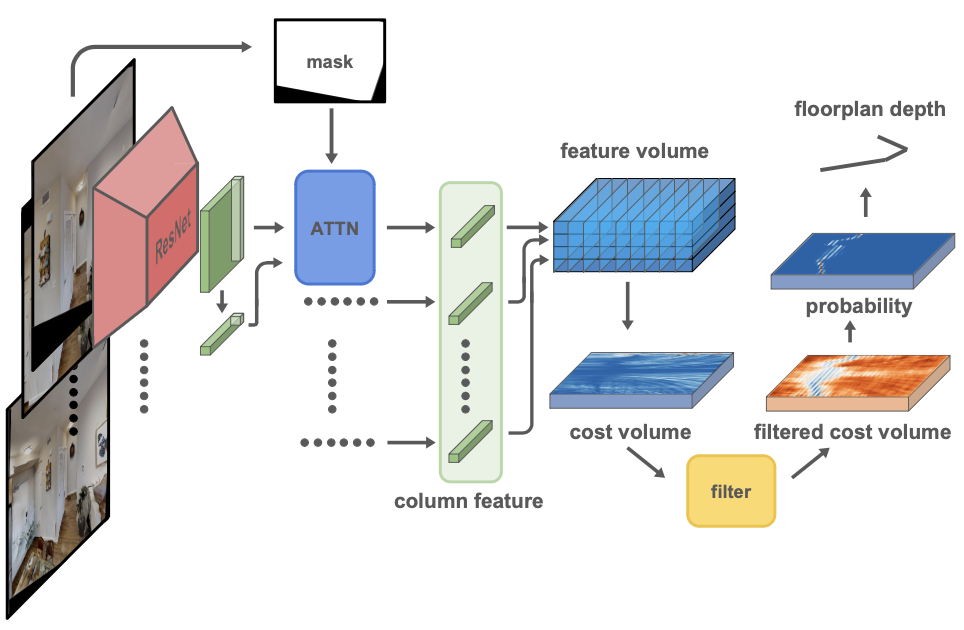}
    \caption{\textbf{Floorplan depth prediction from multiple views.} 
    Column features of the images are extracted and gathered in the reference frame. 
    Their cross-view feature variance is used as cost. 
    A U-Net-like network learns the cost filtering to form a probability distribution, and  
    the floorplan depth is defined by its expectation.}
    \label{fig:multiview}
\end{figure}

\subsection{Learned Complementary Selection}
\label{sec:select}
While monocular depth estimation is independent of camera motion but prone to scale ambiguity, 
Multiview stereo approaches~\cite{mvsnet, deeptam} deliver correct scale, but rely on sufficient baselines and camera overlap. 
Based on these observations, we adopt another MLP that softly selects from the two predictions. 
The network takes the relative poses of the frames and the estimated multiview and monocular mean floorplan depth as inputs and outputs the corresponding weight for the two estimates.
The probability distributions are then fused as the weighted average, \ie,
\begin{equation}
    \mathbf{P}_\textrm{fuse}= w\text{Upsample}(\mathbf{P}_\textrm{mono}) + (1-w)\mathbf{P}_\textrm{mv},
\end{equation}
where $0\leq w\leq1$ is the output by the MLP, $\mathbf{P}_\textrm{mono}$ and $\mathbf{P}_\textrm{mv}$ denote the probability distributions from a single and multi view, respectively. The upsampling ensures the validity of the addition.
The expectation of the fused probability distribution $\mathbf{P}_\textrm{fuse}$ then provides the final depth prediction.

\subsection{Sequential Localization}
\label{sec:histogram}
We use a histogram filter to keep track of the posterior over the entire floorplan. We use the predicted floorplan depth as our observation and the following observation model
\begin{equation}
    p(\mathbf{s}_t|\mathbf{o}_t) = e^{-\frac{1}{\lambda}\|\hat{\mathbf{r}}-\mathbf{r}_{\mathbf{s}_t}\|_1},
\label{eq:observation}
\end{equation}
where $\mathbf{r}_\mathbf{s_t}$ is the floorplan ray at pose $\mathbf{s}_t$, $\hat{\mathbf{r}}$ is the interpolated ray from the floorplan depth prediction and $\lambda$ a constant factor.  We use the relative pose between frames, \ie, ego-motion as the transition model
\begin{equation}
    \mathbf{s}_{t+1} = \mathbf{t}_t\oplus\mathbf{s}_t + \mathbf{\omega}_t,
\end{equation}
where $\mathbf{t}_t= [t_{x,t}, t_{y,t}, t_{\phi,t}]$  and $\mathbf{\omega}_t=[\omega_{x,t}, \omega_{y,t}, \omega_{\phi,t}]$ are the ego-motion and transition noise at time $t$ , respectively, the operator $\oplus$ applies an ego-motion on a state.  Further assuming the transition noise $\mathbf{\omega}_t$ obeys a Gaussian distribution, the transition probability is expressed as
\begin{equation}
    p(\mathbf{s}_{t+1}|\mathbf{s}_t, \mathbf{t}_t) = e^{-\frac{1}{2}(\mathbf{s}_{t+1} - \mathbf{s}_{t}\oplus\mathbf{t}_t)^\top\Sigma^{-1}(\mathbf{s}_{t+1} - \mathbf{s}_{t}\oplus\mathbf{t}_t)},
\end{equation}
where we model $\Sigma=\textrm{diag}(\sigma_x^2, \sigma_y^2, \sigma_\phi^2)$ as covariance of the Gaussian distribution. Applying Bayes rule yields
\begin{equation}
p(\mathbf{s}_{t+1}|\mathbf{o}_t, \mathbf{t}_t) = \frac{1}{Z}\sum_{\mathbf{s}_t\in\mathcal{S}}{p(\mathbf{s}_{t+1}|\mathbf{s}_t, \mathbf{t}_t)p(\mathbf{s}_{t}|\mathbf{o}_t)},
\end{equation}
where $Z$ is a normalization factor and $\mathcal{S}$ the poses space.

\begin{figure}[t!]
    \centering
    \begin{subfigure}{0.8\linewidth}      \includegraphics[width=\linewidth]{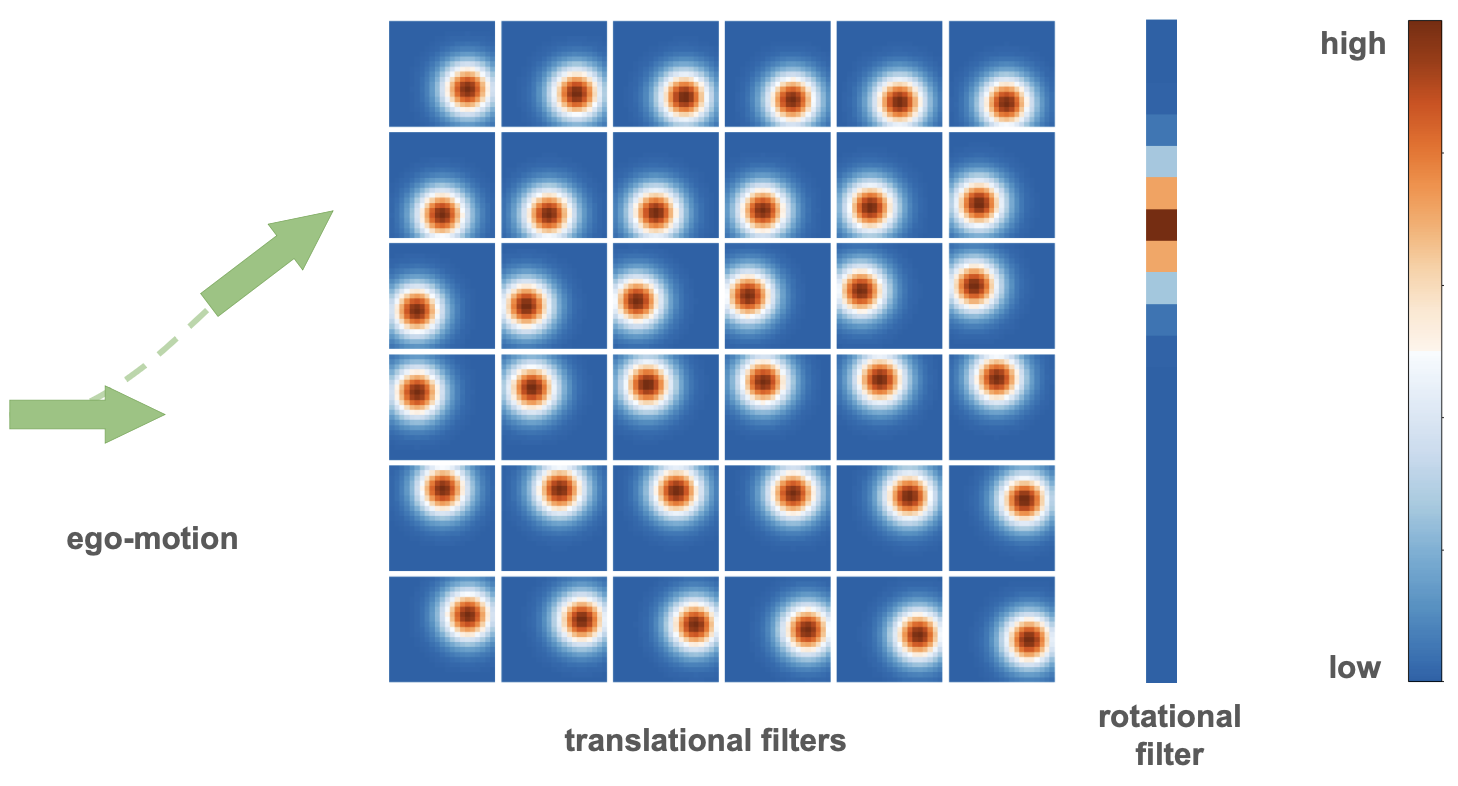}
    \caption{Transitional filters.}
    \end{subfigure}

    \begin{subfigure}{\linewidth}
        \includegraphics[width=\linewidth]{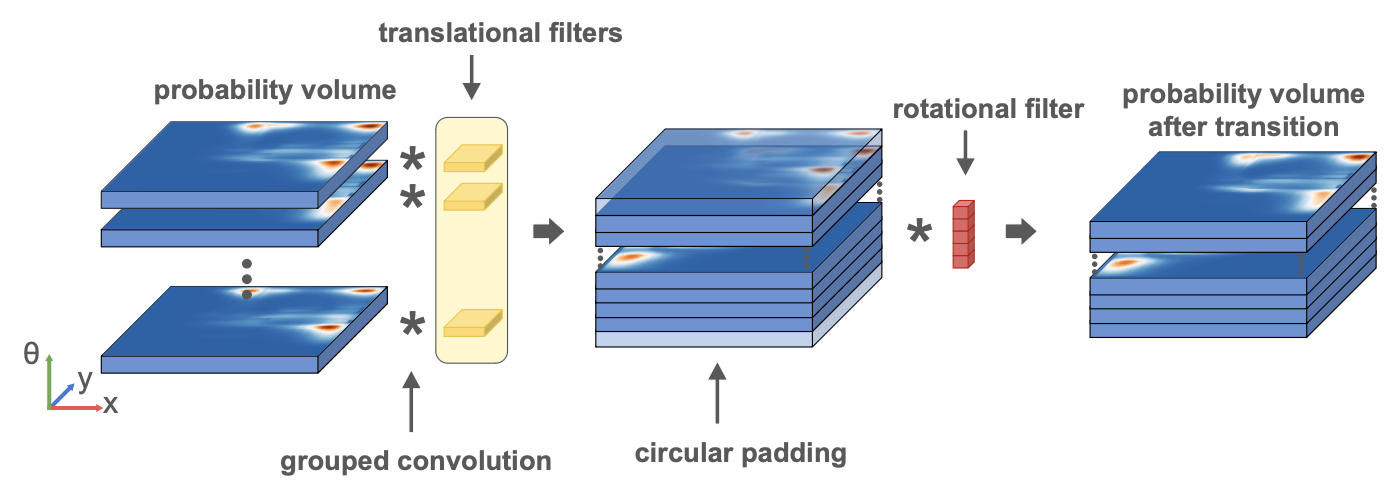}
    \caption{Transition.}
    \end{subfigure}
    
    \caption{\textbf{Transition as grouped convolution.} (a) Illustration of the translational filters (from left to right, top to bottom the filters for 0, 10 to 350$\degree$ ) and the rotational filters derived from a sample ego-motion. (b) The probability volume is divided into $O$ groups, where $O$ is the number of orientations. Each group is convolved with its respective translational filter and stacked back together. After circular padding along the orientation axis, the volume is convolved with the rotational filter to finish the transition step.}
    \label{fig:transition}
\end{figure}

In the following we drop the subscripts indicating the time-step for simplicity.
Our histogram filter represents the posterior as a 3D probability volume containing the probability of being at pose $[s_x, s_y, s_\phi]$. 
The transition is implemented as transition filters~\cite{hist}. 
Unlike previous work~\cite{hist} operating on euclidean $\mathbb{R}^2$ state space, we work on SE2 and transit through ego-motions, so the translation in the world frame depends on the current orientation.
Therefore, we decouple the translation and rotation and apply different 2D translation filters for different orientations, before applying the rotation filter to the entire volume along the orientation axis. 
The 2D translation step can be implemented efficiently as a grouped convolution~\cite{GroupedConvolution}, where each orientation is a group as illustrated in~\cref{fig:transition}. 
The translational filter $\mathbf{T}_\phi$ for orientation $\phi$ can be computed through
\begin{equation}
    \mathbf{T}_\phi(x, y) = e^{-\frac{1}{2}\delta\mathbf{t}^\top \textrm{diag}(\sigma_x^2, \sigma_y^2)^{-1}\delta\mathbf{t}},
\end{equation} 
where
\begin{equation}
    \delta\mathbf{t}=\mathbf{R}_{\phi}^{-1}[x, y]^\top-[t_x, t_y]^\top
\end{equation}
with $\mathbf{R}_{\phi}\in\mathbb{R}^{2\times 2}$ being the rotation matrix with angle ${\phi}$. 
The rotational filter $\mathbf{r}$ is 
\begin{equation}
    \mathbf{r}(\phi) = e^{-\frac{1}{2}(\phi- t_\phi)^2/\sigma_\phi^2}.
\end{equation}
The pose posterior corresponds to the filtered probability volume and we can obtain the (best) pose prediction and its uncertainty by a lookup. 

\section{Training}
\subsection{Dataset}
We collect a customized dataset with perspective images of  108$\degree$ horizontal field of view in iGibson~\cite{igibson}, an indoor simulation environment,  and manually label the floorplans (see~\cref{fig:obseravtion}) from the provided mesh. The dataset consists of 118 distinct indoor environments and is partitioned into training (100), validation (9), and test (9) sets. We collected three datasets according to the type of motions designed to be typical trajectories for a human holding a phone. One including in-place turning, which we refer to as Gibson(g) for general motions, containing 49558 pieces of 4 sequential views, one without (Gibson(f) for forward motions), containing  24779 pieces of 4 sequential views, and one containing 118 pieces of 280 to 5152 steps long trajectories(Gibson(t) for trajectories). We also evaluate the proposed single frame localization on Structured3D~\cite{s3d}, a photorealistic dataset containing 3296 fully furnished indoor environments with in total  78453 perspective images with 80$\degree$ horizontal field of view. For Structured3D, we follow the official split.

\subsection{Virtual Roll Pitch Augmentation}
To cope with non-upright camera poses we propose an augmentation technique during the training through virtual roll pitch angle simulation.
Perspective images with the same principle point and different viewing angles relate to each other through a simple homography as shown in \cref{fig:vrp}.
With known camera intrinsic matrix $\mathbf{K}$, camera roll and pitch angle $\psi$ , $\theta$, the homography from the  original image to the gravity-aligned image is   
\begin{equation}
    \hat{\mathbf{p}} = \mathbf{KR}\mathbf{K}^{-1}\mathbf{p},
    \label{eq:vrp}
\end{equation}
where $\mathbf{p}, \hat{\mathbf{p}}$ are the homogeneous image coordinates of the original pixel and the corresponding pixel in the gravity-aligned image, $\mathbf{R}$ is the rotation matrix to the gravity-aligned pose.
To simulate the virtual roll pitch angle,  we use this to calculate which pixel is observable at angle  $\psi$ , $\theta$ and mask out the unobservable ones. This is equivalent to the gravity-alignment of the image taken at angle  $\psi$ , $\theta$.

\subsection{Training Scheme}

Details on the training procedure can be found in the supplementary material. For all training, we optimize the L1 loss to the ground truth floorplan, except for the monocular network, for which we added a shape loss computed as the cosine similarity, \ie,
 \begin{equation}
    \mathcal{L} = ||\mathbf{d}, \mathbf{d^*}||_{1} + \lambda \frac{\mathbf{d}^\top \mathbf{d^*}}{\max\{||\mathbf{d}||_{2} ||\mathbf{d^*}||_{2}, \epsilon\}},
\label{eq:loss}
\end{equation}
where $\mathbf{d}, \mathbf{d^*}$ are the predicted and the ground truth depth and $\epsilon$ a small constant to prevent from division by zero.
%

\begin{table*}[bp]
\centering
\begin{minipage}{0.6\linewidth}
    \centering\small{
    \begin{tabular}{c c c c c c c c c} 
     \toprule
     & \multicolumn{4}{c}{\textbf{Gibson(f)}} & \multicolumn{4}{c}{\textbf{Structured3D}}\\
     \cmidrule(lr){2-5} \cmidrule(lr){6-9} 
     R@ & 0.1m & 0.5m & 1m & 1m30\degree &0.1m & 0.5m & 1m & 1m30\degree \\ [0.5ex] 
    \midrule
     PF-net & 0 & 2.0 & 6.9 & 1.2 & 0.2 & 1.3 & 3.2 & 0.9 \\ 
     LASER & 0.4 & 6.7 & 13.0 & 10.4 & 0.7 & 6.4 & 10.4 & 8.7 \\
     $\text{Ours}_s$ & \textbf{4.7} & \textbf{28.6} & \textbf{36.6} & \textbf{35.1} & \textbf{1.5} & \textbf{14.6} & \textbf{22.4} & \textbf{21.3} \\ 
     $\text{Ours}_m$ & \textbf{13.2} & \textbf{40.9} & \textbf{45.2} & \textbf{43.7} & - & - & - & -\\ 
     \bottomrule
     \end{tabular}}
     \caption{\textbf{Comparison between our observation model and the baselines.}}
     \label{tab:single}
\end{minipage}\hspace{2mm}
   \begin{minipage}{0.35\linewidth}
   \centering\small{
   \begin{tabular}{c c c c c} 
    \toprule
    & \multicolumn{4}{c}{\textbf{Gibson(g)}}\\
    \cmidrule(lr){2-5}
     R@ & 0.1m & 0.5m & 1m & 1m30\degree \\ 
     \midrule
     PF-net & 1.0 & 1.9 & 5.6 & 1.9  \\ 
     LASER & 0.7& 7.0& 11.8& 9.5\\
     $\text{Ours}_s$  & 4.3& 26.7& 33.7& 32.3\\
     $\text{Ours}_m$ & 9.3& 27.0& 31.0& 29.2\\
     $\text{Ours}_t$ & 10.5& 34.3& 39.6& 38.0\\
     $\text{Ours}_f$  & \textbf{12.2}& \textbf{39.4}& \textbf{44.5}& \textbf{43.2}\\
     \bottomrule
     \end{tabular}
    \caption{\textbf{Complement single and multiview.}}
    \label{tab:general}}
   \end{minipage}
\end{table*}

\section{Results}
We compare our method with the state-of-the-art floorplan localization methods PF-net~\cite{pfnet} and LASER~\cite{laser}, both without semantic labels. 
We sample pose position and orientation at a resolution of 0.1m$\times$0.1m and 10$\degree$.
\subsection{Observation Model}
\begin{figure}[t]
    \centering
    \includegraphics[width=\linewidth]{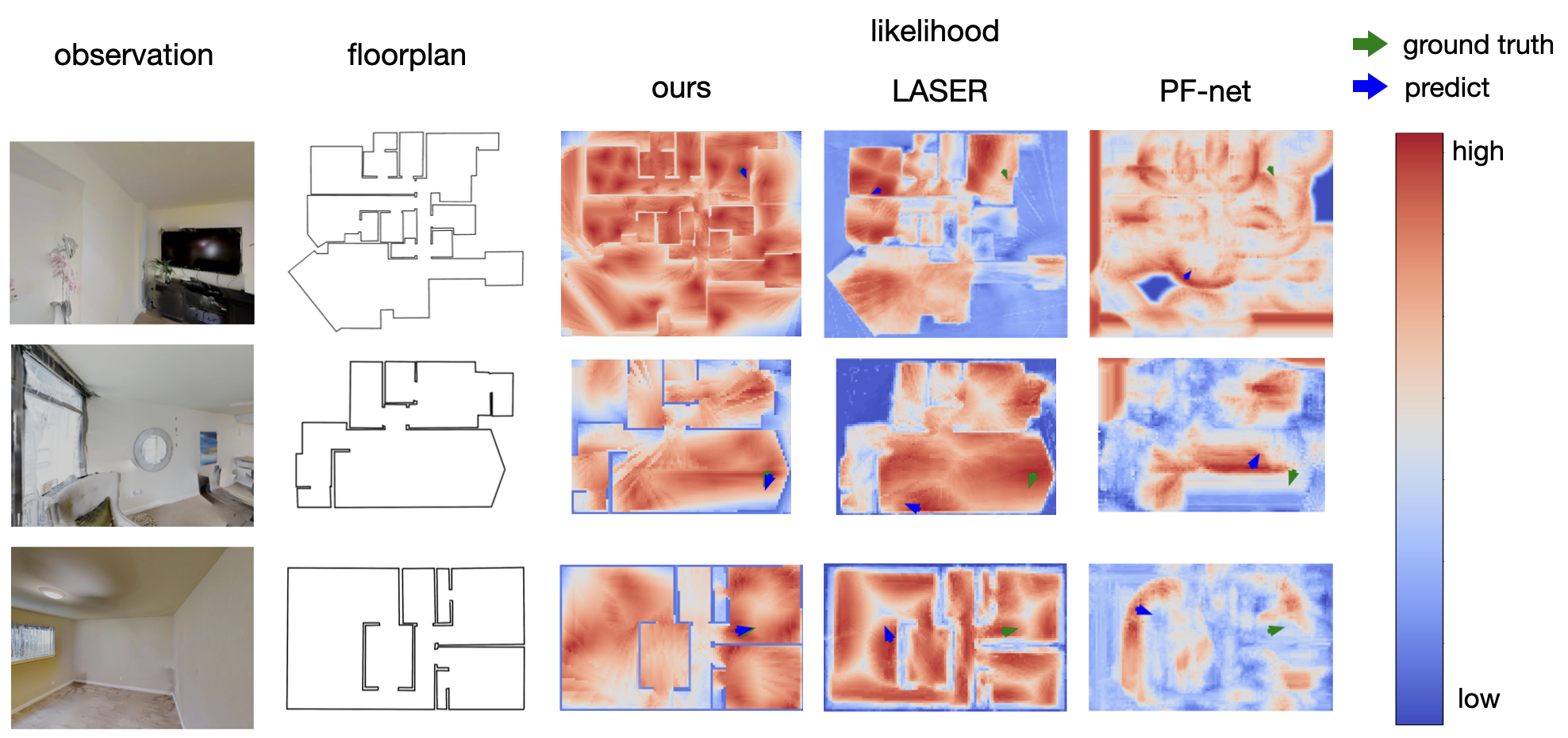}
    \caption{\textbf{Single observation likelihood.} Utilizing the front-facing map patch, PF-net does not account for occlusion or the camera's field of view. Using a set of point features, LASER is not invariant to rotation and translation. Contrary, our 1D ray-scan representation possess such invariance and inherently considers occlusions.}
    \label{fig:obseravtion}
\end{figure}

\cref{fig:obseravtion} provides a qualitative comparison between the methods.  
While all predictions possess multiple modes, our probability estimate appears more accurate, due the accurate floorplan depth estimation and the invariance of the ray representation. 
In the following, we thoroughly investigate the performance of the proposed observation model. 

\noindent
\textbf{Single Frame}.  We evaluate single frame localization accuracy on Gibson(f) and Structured3D. As shown in~\cref{tab:single}, the proposed monocular network, $\text{Ours}_s$ significantly outperforms both baselines on Gibson(f) seeing almost 200\% improvement across all metrics. Also on Structured3D our method surpasses the state of the art by a large margin.
When taking the orientation into account, the recall does not drop much (35.1\% at 1m 30 deg compared to 36.6\% at 1m on Gibson(f) and 21.3\% to 22.4\% on Structured3D). This underlines the accurate orientation estimation of our method. We notice here the performance of LASER on Structured3D does not align with that reported in~\cite{laser}, we suspect this is due to the difference in the dataset (\emph{perspective Structured3D} compared to their perspective images cropped from \emph{panoramic Structured3D}) and the random roll pitch angles this dataset contains.

\noindent\textbf{Multiview}. 
Because the existing indoor datasets either do not provide sequential images or a floorplan, we evaluate the proposed multiview module, $\text{Ours}_m$  only on the collected Gibson dataset. 
\cref{tab:single} verifies that the multiview module can clearly outperform the two baselines on Gibson(f), and notably increases the recall by more than 20\% at all thresholds compared to our monocular module. 
This shows the effectiveness of using multiview geometry cues in the observation model for floorplan depth estimation. 

However, multiview estimation can suffer from small baselines or insufficient overlap present in the general motion dataset, Gibson(g), which includes nearly in-place rotation. 
Here, as shown in~\cref{tab:general}, the recall of the multiview module falls below that of the monocular module for both larger thresholds, 1m and 1m 30$\degree$. 

\begin{figure}[t]
    \centering
    \includegraphics[width=\linewidth,]{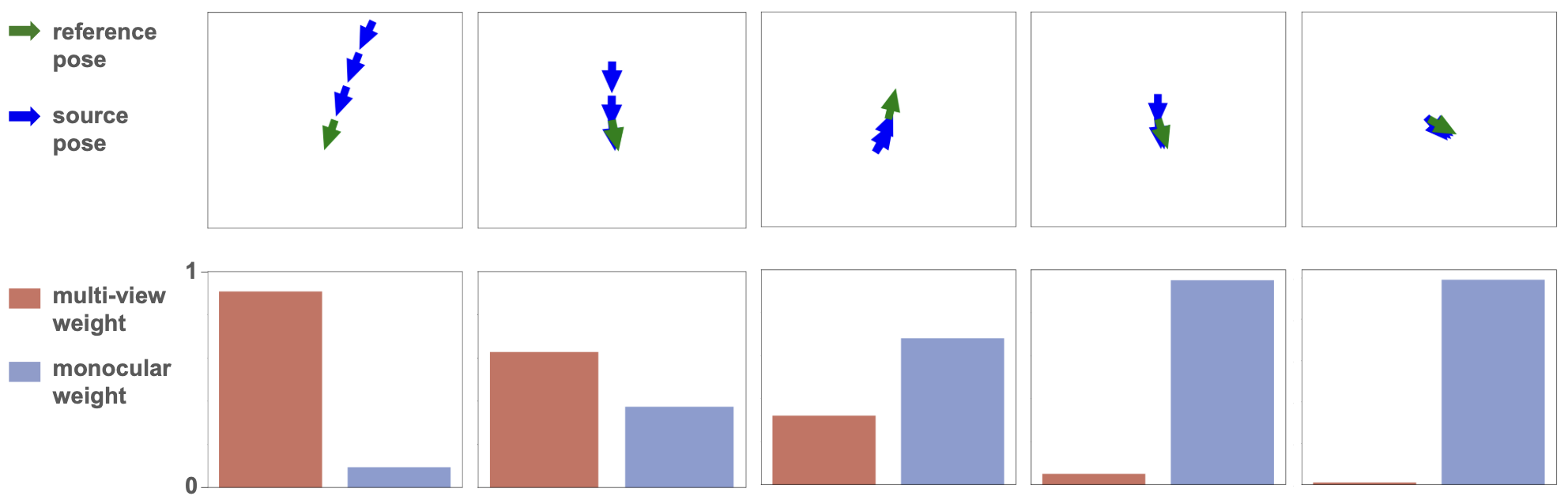}
    \caption{\textbf{Selection weights VS relative poses.} The section network computes a weighted combination of monocular and multi-view observation estimate. The more degenerate the relative poses become the more the monocular estimate is preferred.}
    \label{fig:select}
    \vspace{-1.8mm}
\end{figure}

\noindent
\textbf{The selection network} is evaluated on the general motion dataset Gibson(g) in \cref{tab:general}.
The selection network, $\text{Ours}_f$, delivers a 30-50$\%$ improvement across all precisions compared to both individual networks $\text{Ours}_s$ and $\text{Ours}_m$.
As a baseline selection we also evaluate selection by thresholding the relative motions between sequential frames named $\text{Ours}_t$.
While this baseline achieves a $18\%$ improvement over the individual networks, which further proves the idea of complementing monocular with multiview estimation, 
the selection network learns a more sophisticated selection rule and achieves at least an additional extra 12$\%$ improvement. 
Examples of the selection decisions are illustrated in~\cref{fig:select}.

\noindent
\begin{figure}[t]
    \centering
    \begin{subfigure}{0.32\linewidth}
    \centering
    \includegraphics[width=\linewidth]{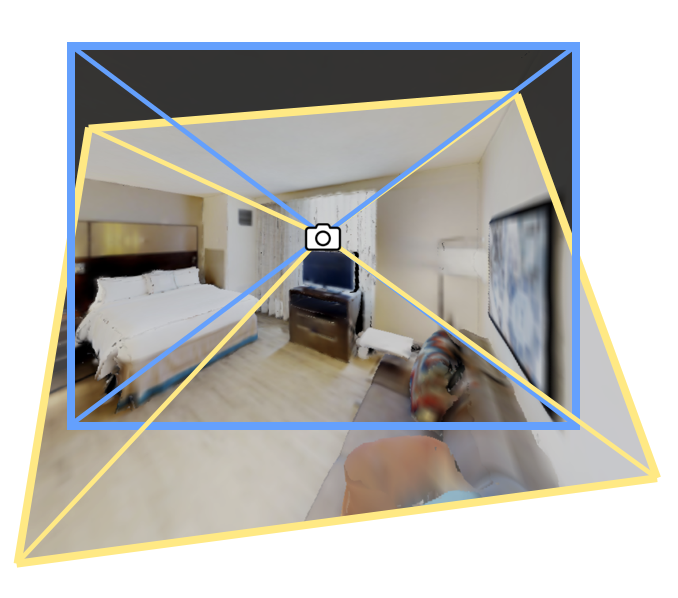}
    \caption{Virtual roll pitch.}
    \label{fig:vrp}
    \end{subfigure}
    \begin{subfigure}{0.66\linewidth}
        \includegraphics[width=\linewidth]{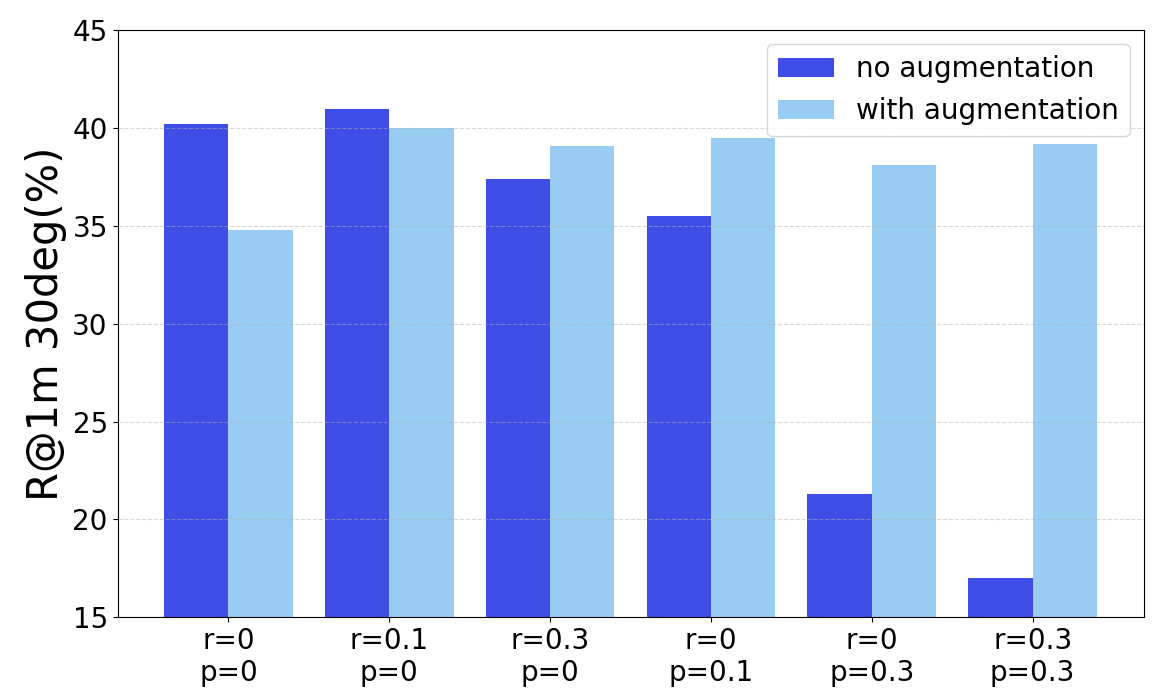}
    \caption{Robust to non-upright camera poses.}\label{fig:robust}
    \end{subfigure}
    \caption{\textbf{Virtual roll pitch augmentation.} (a) After gravity alignment we mask out unobservable pixels (in black). During training we augment the data accordingly. (b) If trained without augmentation, the recall of the network decreases as the roll and pitch angle increases. Training with augmentation significantly increases robustness against non-upright camera poses.  
    }
\end{figure}
%
%
\textbf{Virtual Roll-Pitch.} 
\cref{fig:robust} compares the recall of the monocular module trained with and without virtual roll-pitch augmentation on Gibson(f) at 1m$\times$1m$\times$30$\degree$ resolution. 
The network trained without augmentation shows decreasing recall when the roll pitch disturbances are imposed (especially for large pitch angles).
In contrast the proposed virtual roll-pitch augmentation increases the robustness of the recall against non-upright poses. 

\subsection{Sequential localization}
\begin{figure}[t]
    \centering
    \includegraphics[width=0.975\linewidth]{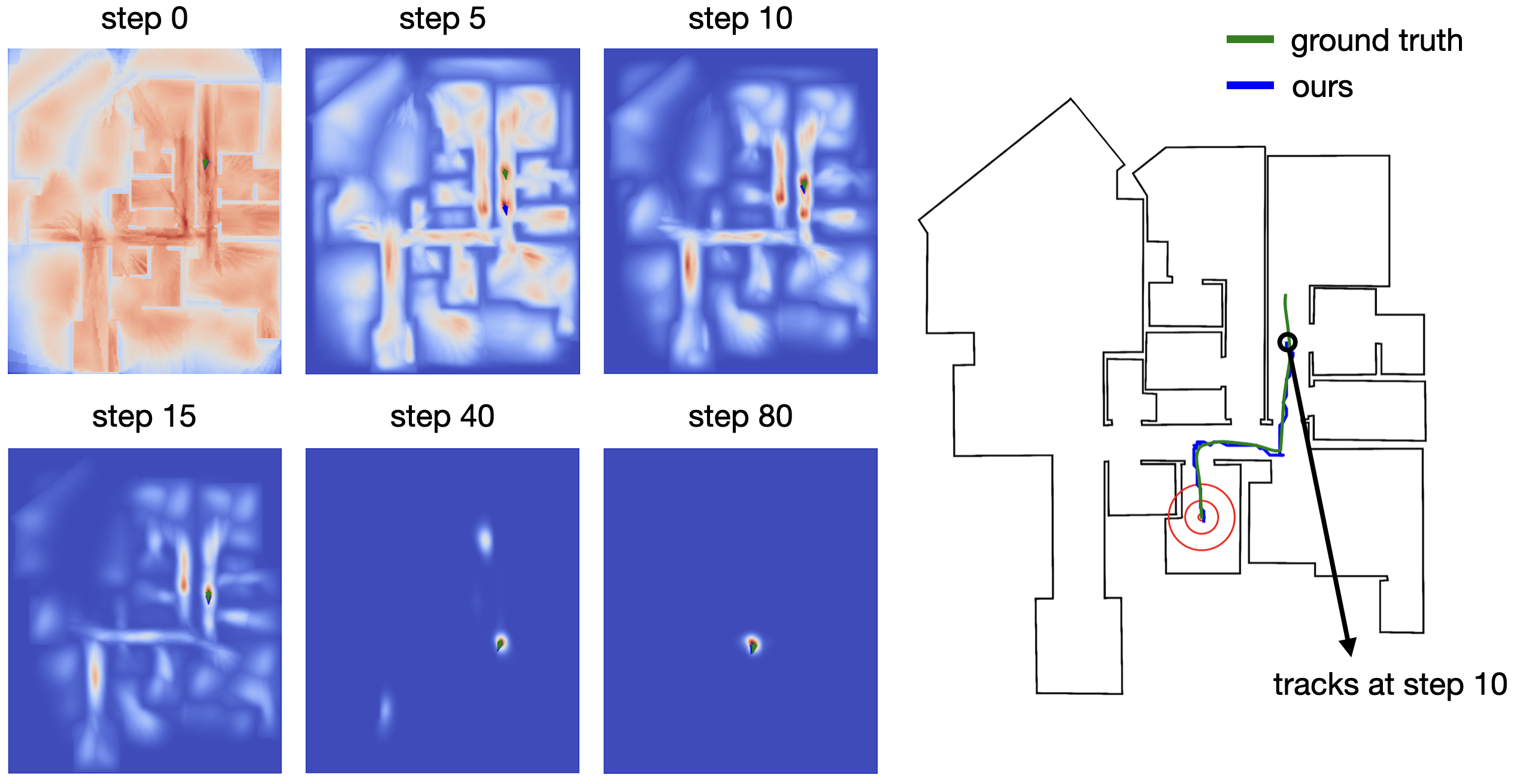}
    \caption{\textbf{Posterior evolution and trajectory.} Our strong observation model already provides an accurate estimation at the initial step. Due to the ambiguous nature of the floorplan (the hallway), the posterior estimates shows multi-modality. After 10 steps, our system tracks firmly at a frequency of 27Hz on this 18.4m$\times$15.5m floorplan using a laptop NVidia RTX 3070Ti GPU.}
    \label{fig:post}
    \vspace{-1.5mm} 
\end{figure}

Our full sequential localization pipeline is evaluated on the Gibson(t) dataset, containing long simulated trajectories.
A qualitative study in~\cref{fig:post} shows that the proposed histogram filter can effectively maintain a global posterior of the camera pose. 
At the start the distribution has multiple modes, as the camera movement provides more and more evidence, the distribution converges to a single sharp peak.

\begin{figure}[t]
    \centering
    \includegraphics[width=0.7\linewidth]{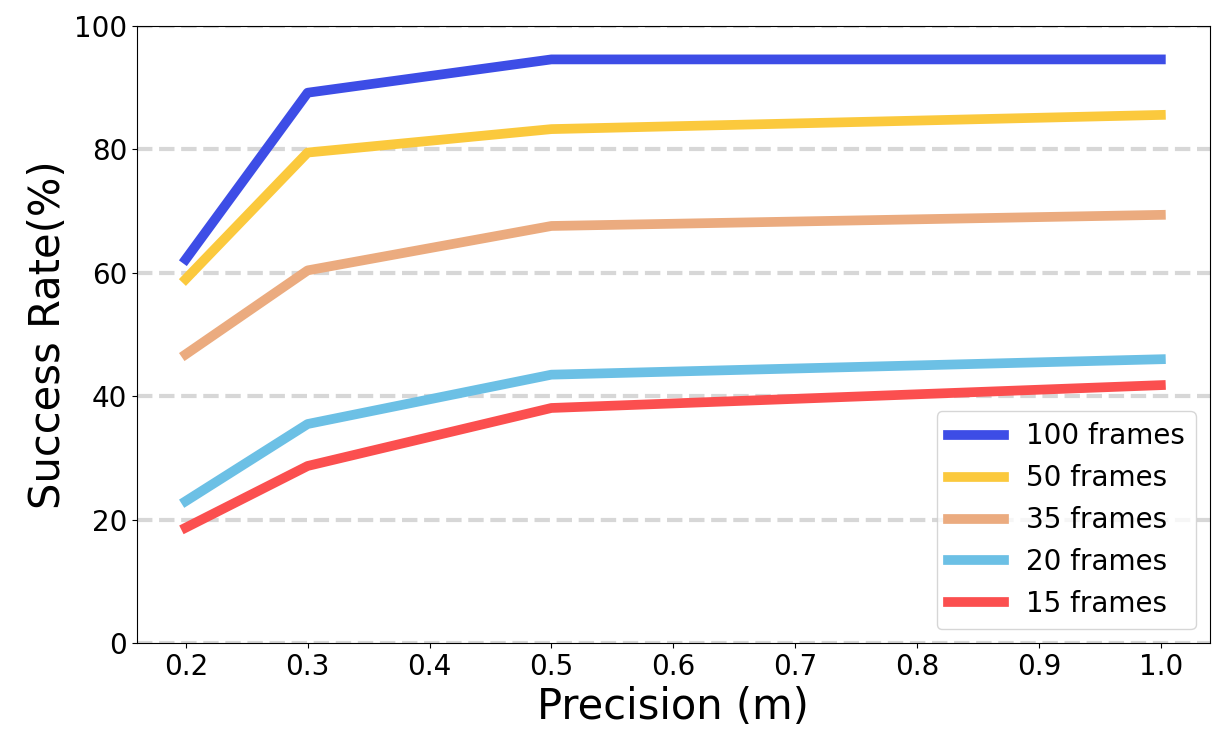}
    \caption{\textbf{Localization success rate vs. precision threshold for different filter history sizes}. The more frames are used within the filter, the higher the localization success rate.}
    \label{fig:success/frame}
    \vspace{-4mm} 
\end{figure}

\begin{table}[b!]\centering 
{\small 
\begin{tabular}{c  c c c} 
 \toprule
 &LASER & $\text{Ours}_s$  & $\text{Ours}_f$\\
 \midrule
{\small Success rate@1m ($\%$)} & 59.5 & 89.2 &\textbf{ 94.6} \\ 
{\small  RMSE(succeeded) (m)}& 0.39 & 0.18 & \textbf{0.12} \\ 
{\small  RMSE(all) (m)} & 1.96 & 0.88 & \textbf{0.51} \\ 
 \bottomrule
 \end{tabular}
 }
\caption{\textbf{Comparison of observation models integrated into our histogram filter.} RMSEs are computed from the last 10 frames}
\label{tab:seq}
\end{table}
\noindent
\textbf{Success Rate}. 
As a metric we consider sequential localization at Xm as successful if the prediction stays within a radius of Xm over the last 10 frames. 
We integrate the baseline observation models into our histogram filter and compare them against our pipeline in \cref{tab:seq}. 
Our full system achieves a success rate of 94.6\% at 1m using a history of 100 frames, surpassing the two baselines by more than 58\%, and our monocular observation by 10\%. 
We also compare the RMSE (over the last 10 frames) of our trajectory tracking in both \textit{succeeded} and \textit{all} runs. 
Here, our full pipeline delivers 70\% lower error (0.12 m and 0.51 m, respectively) than the baselines. 
We compare success rates for various number of frames in~\cref{fig:success/frame}. In general, using more frames increases the success rates.

\noindent
\textbf{Timing.}
\cref{tab:timing} compares the runtime of different combinations of observation and filtering models. Despite the slightly slower feature extraction of our proposed observation model, the rapid matching  helps it to achieve the highest iteration rates.
The particle filter (PF) suffers from expensive resampling and feature rendering and demands instanciating a large number of samples for global localization in a large area. 
Analogously, our histogram filter (HF) utilizes presampled "particles", constructed offline, and can avoid constant rerendering at runtime. 
As a result, our histogram filter achieves 45\% faster iteration than the particle filter.
\begin{table}[b!]
\centering{\small
 \begin{tabular}{c c c c c} 
 \toprule
  &\multirow{2}{*}{\parbox{1.5 cm}{\centering Feature \mbox{Extraction(s)}}} & \multirow{2}{*}{Matching(s)} &\multicolumn{2}{c}{Iteration} \\
 \cmidrule(lr){4-4} \cmidrule(lr){5-5}
  & & & HF& PF\\
 \midrule
 PF-net(obs) & 0.042 & 2.375 & - & -\\ 
 LASER(obs) & 0.008 & 0.224 & 0.241 & 0.287\\ 
  $\text{Ours}_f$ & 0.033 & 0.003 & 0.037 & 0.067\\
 \bottomrule
 \end{tabular}}
\caption{\textbf{Timing.} Because PF-net is too slow we do not test its performance in filters.}
\label{tab:timing}
\end{table}



\section{Real-world Experiment}

\begin{figure}[t]
    \centering
    \includegraphics[width=0.86\linewidth]{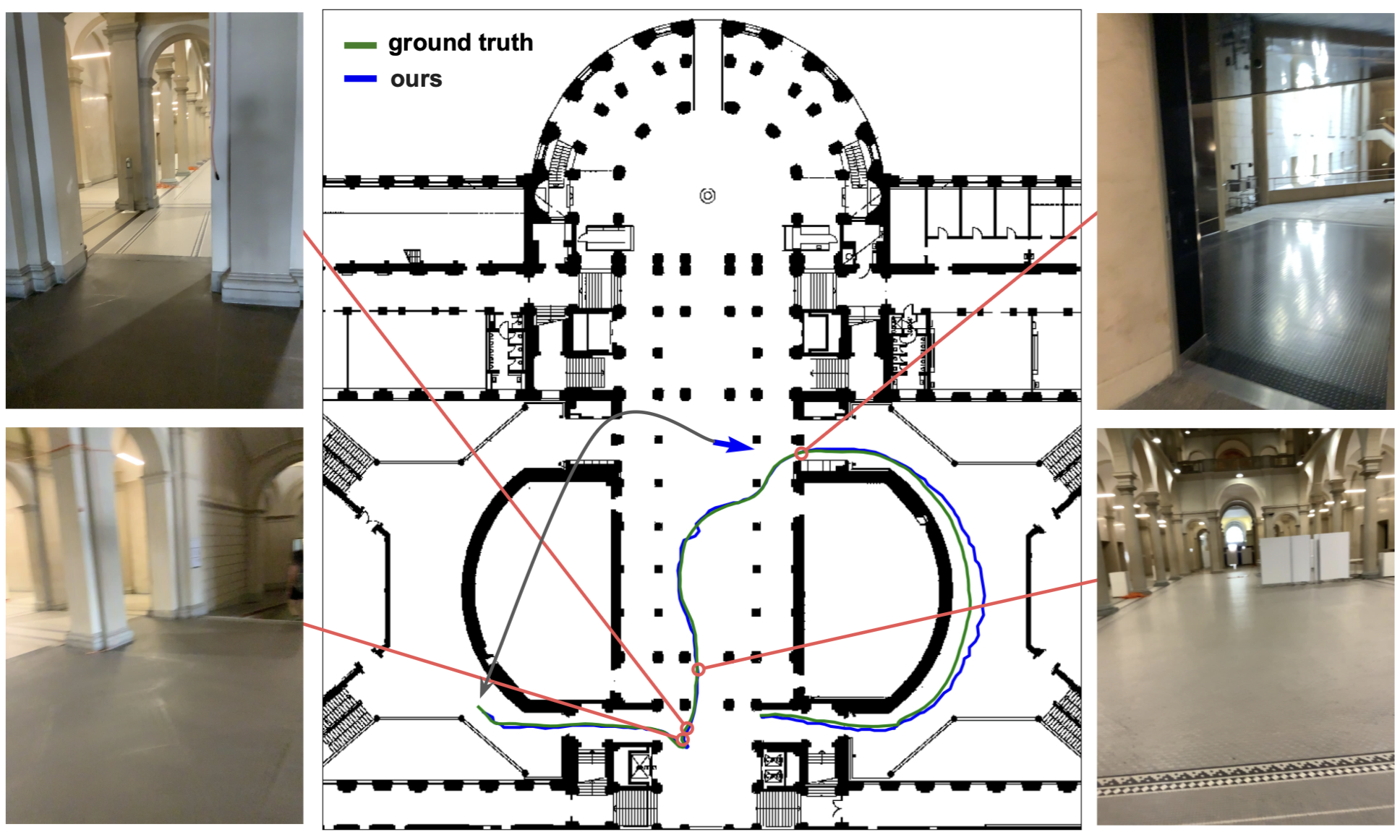}
    \caption{\textbf{Sequential localization in HGE.}  The localization area is 75m$\times$81m with challenging observations including motion blur, non-lambertian surfaces, ambiguities and occlusions. Our trajectory tracks the ground truth closely from the second step. It deviates slightly later due to the ambiguous floorplan labeling, however, recovers shortly thanks to the filters and converges to a sharp posterior estimation in the end.}
    \label{fig:lamar}
    \vspace{-4mm} 
\end{figure}

Since no real-world indoor dataset with both sequential observations and floorplan exist that allows training and testing, we show the potential of our pipeline in real-world scenario by customizing LaMAR~\cite{lamar}. LaMAR is a real world dataset containing three scenes. We select the trajectories in HGE indoor scene containing trajectories within a single floor, and split it into train and test set.  
We use our single frame observation model with the proposed histogram filter to localize.
The entire floor has an area of 80m$\times$120m, and the data includes challenging observations as shown in~\cref{fig:lamar}. We use the data within 75m$\times$81m and localize within the floorplan of the same size. Our system localizes and tracks the camera pose from the second step and closely follows it afterwards. Despite the large scene scale, our histogram filter is still efficient enough to localize at 3 hz. 

\section{Limitations and Conclusion}
Through the process, we realized a lack of indoor datasets with sequential observations and floorplan.
Although we tried to mitigate this by collecting a dataset in a simulated environment, more real-world datasets are highly desirable to close the domain gap.
While our proposed system effectively uses geometric cues, ambiguities could be further reduced by utilizing semantic information from both the image and the floorplan.
%
In this work, we present a data-driven and probabilistic model for localization within a floorplan. 
The system is more practical than previous methods, demanding only consumer hardware, 
perspective RGB images and non-upright camera poses, while operating at very high frame-rates.
Our system allows for both accurate single-frame and sequential localization in unvisited environments.
It outperforms the state-of-the-art in both tasks across different datasets and various metrics by a significant margin. 
Finally, we illustrate its real world potential on a challenging large scale indoor dataset. 
Our work could be interesting in many indoor AR/VR applications and boost robot autonomy in indoor environments.

{
    \small
    \bibliographystyle{ieeenat_fullname}
    \bibliography{main}

\begin{thebibliography}{56}
\providecommand{\natexlab}[1]{#1}
\providecommand{\url}[1]{\texttt{#1}}
\expandafter\ifx\csname urlstyle\endcsname\relax
  \providecommand{\doi}[1]{doi: #1}\else
  \providecommand{\doi}{doi: \begingroup \urlstyle{rm}\Url}\fi

\bibitem[Arandjelovic et~al.(2016)Arandjelovic, Gronat, Torii, Pajdla, and Sivic]{netvlad}
Relja Arandjelovic, Petr Gronat, Akihiko Torii, Tomas Pajdla, and Josef Sivic.
\newblock Netvlad: Cnn architecture for weakly supervised place recognitio.
\newblock In \emph{CVPR}, pages 5297--5307, 2016.

\bibitem[Balntas et~al.(2018)Balntas, Li, and Prisacariu]{relocnet}
Vassileios Balntas, Shuda Li, and Victor Prisacariu.
\newblock Relocnet: Continuous metric learning relocalisation using neural nets.
\newblock In \emph{ECCV}, pages 751--767, 2018.

\bibitem[Boniardi et~al.(2017)Boniardi, Caselitz, K{\"u}mmerle, and Burgard]{lidar2}
Federico Boniardi, Tim Caselitz, Rainer K{\"u}mmerle, and Wolfram Burgard.
\newblock Robust lidar-based localization in architectural floor plans.
\newblock In \emph{IROS}, pages 3318--3324, 2017.

\bibitem[Boniardi et~al.(2019{\natexlab{a}})Boniardi, Caselitz, K{\"u}mmerle, and Burgard]{lidar1}
Federico Boniardi, Tim Caselitz, Rainer K{\"u}mmerle, and Wolfram Burgard.
\newblock A pose graph-based localization system for long-term navigation in cad floor plans.
\newblock pages 84--97, 2019{\natexlab{a}}.

\bibitem[Boniardi et~al.(2019{\natexlab{b}})Boniardi, Valada, Mohan, Caselitz, and Burgard]{edge}
Federico Boniardi, Abhinav Valada, Rohit Mohan, Tim Caselitz, and Wolfram Burgard.
\newblock Robot localization in floor plans using a room layout edge extraction network.
\newblock In \emph{IROS}, pages 5291--5297, 2019{\natexlab{b}}.

\bibitem[Brachmann et~al.(2017)Brachmann, Krull, Nowozin, Shotton, Michel, Gumhold, and Rother]{dsac}
Eric Brachmann, Alexander Krull, Sebastian Nowozin, Jamie Shotton, Frank Michel, Stefan Gumhold, and Carsten Rother.
\newblock Dsac-differentiable ransac for camera localization.
\newblock In \emph{CVPR}, pages 6684--6692, 2017.

\bibitem[Cheng et~al.(2020)Cheng, Xu, Zhu, Li, Li, Ramamoorthi, and Su]{adaptmvs}
Shuo Cheng, Zexiang Xu, Shilin Zhu, Zhuwen Li, Li~Erran Li, Ravi Ramamoorthi, and Hao Su.
\newblock Deep stereo using adaptive thin volume representation with uncertainty awareness.
\newblock In \emph{CVPR}, pages 2524--2534, 2020.

\bibitem[Chu et~al.(2015)Chu, Kim, and Chen]{mimick}
Hang Chu, Dong~Ki Kim, and Tsuhan Chen.
\newblock You are here: Mimicking the human thinking process in reading floor-plans.
\newblock In \emph{ICCV}, pages 2210--2218, 2015.

\bibitem[Collins(1996)]{planesweep}
Robert~T Collins.
\newblock A space-sweep approach to true multi-image matching.
\newblock In \emph{CVPR}, pages 358--363, 1996.

\bibitem[Dellaert et~al.(1999)Dellaert, Fox, Burgard, and Thrun]{mcl}
Frank Dellaert, Dieter Fox, Wolfram Burgard, and Sebastian Thrun.
\newblock Monte carlo localization for mobile robots.
\newblock In \emph{ICRA}, pages 1322--1328, 1999.

\bibitem[Delmerico et~al.(2019)Delmerico, Mintchev, Giusti, Gromov, Melo, Horvat, Cadena, Hutter, Ijspeert, Floreano, et~al.]{rescue}
Jeffrey Delmerico, Stefano Mintchev, Alessandro Giusti, Boris Gromov, Kamilo Melo, Tomislav Horvat, Cesar Cadena, Marco Hutter, Auke Ijspeert, Dario Floreano, et~al.
\newblock The current state and future outlook of rescue robotics.
\newblock \emph{Journal of Field Robotics}, 36\penalty0 (7):\penalty0 1171--1191, 2019.

\bibitem[Eigen and Fergus(2015)]{eigen2}
David Eigen and Rob Fergus.
\newblock Predicting depth, surface normals and semantic labels with a common multi-scale convolutional architecture.
\newblock In \emph{NeurIPS}, pages 2650--2658, 2015.

\bibitem[Eigen et~al.(2014)Eigen, Puhrsch, and Fergus]{eigen1}
David Eigen, Christian Puhrsch, and Rob Fergus.
\newblock Depth map prediction from a single image using a multi-scale deep network.
\newblock 2014.

\bibitem[Godard et~al.(2017)Godard, Mac~Aodha, and Brostow]{leftright}
Cl{\'e}ment Godard, Oisin Mac~Aodha, and Gabriel~J Brostow.
\newblock Unsupervised monocular depth estimation with left-right consistency.
\newblock In \emph{CVPR}, pages 270--279, 2017.

\bibitem[He et~al.(2016)He, Zhang, Ren, and Sun]{resnet}
Kaiming He, Xiangyu Zhang, Shaoqing Ren, and Jian Sun.
\newblock Deep residual learning for image recognition.
\newblock In \emph{CVPR}, pages 770--778, 2016.

\bibitem[Howard-Jenkins and Prisacariu(2022)]{lalaloc++}
Henry Howard-Jenkins and Victor~Adrian Prisacariu.
\newblock Lalaloc++: Global floor plan comprehension for layout localisation in unvisited environments.
\newblock In \emph{ECCV}, pages 693--709, 2022.

\bibitem[Howard-Jenkins et~al.(2021)Howard-Jenkins, Ruiz-Sarmiento, and Prisacariu]{lalaloc}
Henry Howard-Jenkins, Jose-Raul Ruiz-Sarmiento, and Victor~Adrian Prisacariu.
\newblock Lalaloc: Latent layout localisation in dynamic, unvisited environments.
\newblock In \emph{ICCV}, pages 10107--10116, 2021.

\bibitem[Ito et~al.(2014)Ito, Endres, Kuderer, Tipaldi, Stachniss, and Burgard]{wifi}
Seigo Ito, Felix Endres, Markus Kuderer, Gian~Diego Tipaldi, Cyrill Stachniss, and Wolfram Burgard.
\newblock W-rgb-d: floor-plan-based indoor global localization using a depth camera and wifi.
\newblock In \emph{ICRA}, pages 417--422, 2014.

\bibitem[Jonschkowski and Brock(2016)]{hist}
Rico Jonschkowski and Oliver Brock.
\newblock End-to-end learnable histogram filters.
\newblock In \emph{Workshop on Deep Learning for Action and Interaction at NIPS}, 2016.

\bibitem[Karkus et~al.(2018)Karkus, Hsu, and Lee]{pfnet}
Peter Karkus, David Hsu, and Wee~Sun Lee.
\newblock Particle filter networks with application to visual localization.
\newblock In \emph{CoRL}, pages 169--178, 2018.

\bibitem[Kendall et~al.(2015)Kendall, Grimes, and Cipolla]{posenet}
Alex Kendall, Matthew Grimes, and Roberto Cipolla.
\newblock Posenet: A convolutional network for real-time 6-dof camera relocalization.
\newblock In \emph{ICCV}, pages 2938--2946, 2015.

\bibitem[Krizhevsky et~al.(2012)Krizhevsky, Sutskever, and Hinton]{GroupedConvolution}
Alex Krizhevsky, Ilya Sutskever, and Geoffrey~E Hinton.
\newblock Imagenet classification with deep convolutional neural networks.
\newblock In \emph{NeurIPS}, 2012.

\bibitem[Li et~al.()Li, Ang, and Rus]{lidar4}
Zhikai Li, Marcelo~H Ang, and Daniela Rus.
\newblock Online localization with imprecise floor space maps using stochastic gradient descent.
\newblock In \emph{IROS}, pages 8571--8578.

\bibitem[Liu et~al.(2017)Liu, Li, and Dai]{2d3d4}
Liu Liu, Hongdong Li, and Yuchao Dai.
\newblock Efficient global 2d-3d matching for camera localization in a large-scale 3d map.
\newblock In \emph{ICCV}, pages 2372--2381, 2017.

\bibitem[Luo et~al.(2019)Luo, Guan, Ju, Huang, and Luo]{pmvsnet}
Keyang Luo, Tao Guan, Lili Ju, Haipeng Huang, and Yawei Luo.
\newblock P-mvsnet: Learning patch-wise matching confidence aggregation for multi-view stereo.
\newblock In \emph{ICCV}, pages 10452--10461, 2019.

\bibitem[Luo et~al.(2020)Luo, Guan, Ju, Wang, Chen, and Luo]{attnmvs}
Keyang Luo, Tao Guan, Lili Ju, Yuesong Wang, Zhuo Chen, and Yawei Luo.
\newblock Attention-aware multi-view stereo.
\newblock In \emph{CVPR}, pages 1590--1599, 2020.

\bibitem[Mendez et~al.(2020)Mendez, Hadfield, Pugeault, and Bowden]{sedar}
Oscar Mendez, Simon Hadfield, Nicolas Pugeault, and Richard Bowden.
\newblock Sedar: Reading floorplans like a human—using deep learning to enable human-inspired localisation.
\newblock \emph{IJCV}, 128:\penalty0 1286--1310, 2020.

\bibitem[Min et~al.(2022)Min, Khosravan, Bessinger, Narayana, Kang, Dunn, and Boyadzhiev]{laser}
Zhixiang Min, Naji Khosravan, Zachary Bessinger, Manjunath Narayana, Sing~Bing Kang, Enrique Dunn, and Ivaylo Boyadzhiev.
\newblock Laser: Latent space rendering for 2d visual localization.
\newblock In \emph{CVPR}, pages 11122--11131, 2022.

\bibitem[Peng et~al.(2022)Peng, Wang, Wang, Lai, and Wang]{unimvs}
Rui Peng, Rongjie Wang, Zhenyu Wang, Yawen Lai, and Ronggang Wang.
\newblock Rethinking depth estimation for multi-view stereo: A unified representation.
\newblock In \emph{CVPR}, pages 8645--8654, 2022.

\bibitem[Qi et~al.(2017)Qi, Su, Mo, and Guibas]{pointnet}
Charles~R Qi, Hao Su, Kaichun Mo, and Leonidas~J Guibas.
\newblock Point{N}et: Deep learning on point sets for 3d classification and segmentation.
\newblock In \emph{CVPR}, pages 652--660, 2017.

\bibitem[Ranftl et~al.(2020)Ranftl, Lasinger, Hafner, Schindler, and Koltun]{midas}
Ren{\'e} Ranftl, Katrin Lasinger, David Hafner, Konrad Schindler, and Vladlen Koltun.
\newblock Towards robust monocular depth estimation: Mixing datasets for zero-shot cross-dataset transfer.
\newblock \emph{IEEE TPAMI}, 44\penalty0 (3):\penalty0 1623--1637, 2020.

\bibitem[Ranftl et~al.(2021)Ranftl, Bochkovskiy, and Koltun]{vitmono}
Ren{\'e} Ranftl, Alexey Bochkovskiy, and Vladlen Koltun.
\newblock Vision transformers for dense prediction.
\newblock In \emph{ICCV}, pages 12179--12188, 2021.

\bibitem[Samano et~al.(2020)Samano, Zhou, and Calway]{geolocation}
Noe Samano, Mengjie Zhou, and Andrew Calway.
\newblock You are here: Geolocation by embedding maps and images.
\newblock In \emph{ECCV}, pages 502--518, 2020.

\bibitem[Sarlin et~al.(2019)Sarlin, Cadena, Siegwart, and Dymczyk]{hloc}
Paul-Edouard Sarlin, Cesar Cadena, Roland Siegwart, and Marcin Dymczyk.
\newblock From coarse to fine: Robust hierarchical localization at large scale.
\newblock In \emph{CVPR}, pages 12716--12725, 2019.

\bibitem[Sarlin et~al.(2022)Sarlin, Dusmanu, Sch{\"o}nberger, Speciale, Gruber, Larsson, Miksik, and Pollefeys]{lamar}
Paul-Edouard Sarlin, Mihai Dusmanu, Johannes~L Sch{\"o}nberger, Pablo Speciale, Lukas Gruber, Viktor Larsson, Ondrej Miksik, and Marc Pollefeys.
\newblock Lamar: Benchmarking localization and mapping for augmented reality.
\newblock In \emph{ECCV}, pages 686--704, 2022.

\bibitem[Sarlin et~al.(2023)Sarlin, DeTone, Yang, Avetisyan, Straub, Malisiewicz, Bul{\`o}, Newcombe, Kontschieder, and Balntas]{orienter}
Paul-Edouard Sarlin, Daniel DeTone, Tsun-Yi Yang, Armen Avetisyan, Julian Straub, Tomasz Malisiewicz, Samuel~Rota Bul{\`o}, Richard Newcombe, Peter Kontschieder, and Vasileios Balntas.
\newblock Orienternet: Visual localization in 2d public maps with neural matching.
\newblock In \emph{CVPR}, pages 21632--21642, 2023.

\bibitem[Sattler et~al.(2011)Sattler, Leibe, and Kobbelt]{2d3d1}
Torsten Sattler, Bastian Leibe, and Leif Kobbelt.
\newblock Fast image-based localization using direct 2d-to-3d matching.
\newblock In \emph{ICCV}, pages 667--674, 2011.

\bibitem[Sattler et~al.(2012)Sattler, Leibe, and Kobbelt]{2d3d2}
Torsten Sattler, Bastian Leibe, and Leif Kobbelt.
\newblock Improving image-based localization by active correspondence search.
\newblock In \emph{ECCV}, pages 752--765, 2012.

\bibitem[Sattler et~al.(2016)Sattler, Leibe, and Kobbelt]{2d3d3}
Torsten Sattler, Bastian Leibe, and Leif Kobbelt.
\newblock Efficient \& effective prioritized matching for large-scale image-based localization.
\newblock \emph{PAMI}, 39\penalty0 (9):\penalty0 1744--1756, 2016.

\bibitem[Schindler et~al.(2007)Schindler, Brown, and Szeliski]{cityscale}
Grant Schindler, Matthew Brown, and Richard Szeliski.
\newblock City-scale location recognition.
\newblock In \emph{CVPR}, pages 1--7, 2007.

\bibitem[Shen et~al.()Shen, Xia, Li, Mart{\'\i}n-Mart{\'\i}n, Fan, Wang, P{\'e}rez-D’Arpino, Buch, Srivastava, Tchapmi, et~al.]{igibson}
Bokui Shen, Fei Xia, Chengshu Li, Roberto Mart{\'\i}n-Mart{\'\i}n, Linxi Fan, Guanzhi Wang, Claudia P{\'e}rez-D’Arpino, Shyamal Buch, Sanjana Srivastava, Lyne Tchapmi, et~al.
\newblock igibson 1.0: A simulation environment for interactive tasks in large realistic scenes.
\newblock In \emph{IROS}.

\bibitem[Shotton et~al.(2013)Shotton, Glocker, Zach, Izadi, Criminisi, and Fitzgibbon]{forest}
Jamie Shotton, Ben Glocker, Christopher Zach, Shahram Izadi, Antonio Criminisi, and Andrew Fitzgibbon.
\newblock Scene coordinate regression forests for camera relocalization in rgb-d images.
\newblock In \emph{CVPR}, pages 2930--2937, 2013.

\bibitem[Teed and Deng(2020)]{deepv2d}
Zachary Teed and Jia Deng.
\newblock Deep{V}2{D}: Video to depth with differentiable structure from motion.
\newblock In \emph{ICLR}, 2020.

\bibitem[Valentin et~al.(2015)Valentin, Nie{\ss}ner, Shotton, Fitzgibbon, Izadi, and Torr]{forest1}
Julien Valentin, Matthias Nie{\ss}ner, Jamie Shotton, Andrew Fitzgibbon, Shahram Izadi, and Philip~HS Torr.
\newblock Exploiting uncertainty in regression forests for accurate camera relocalization.
\newblock In \emph{CVPR}, pages 4400--4408, 2015.

\bibitem[van~der Merwe et~al.(2000)van~der Merwe, Doucet, de~Freitas, and Wan]{particleFilter}
Rudolph van~der Merwe, Arnaud Doucet, Nando de Freitas, and Eric Wan.
\newblock The unscented particle filter.
\newblock In \emph{NeurIPS}, 2000.

\bibitem[Vaswani et~al.(2017)Vaswani, Shazeer, Parmar, Uszkoreit, Jones, Gomez, Kaiser, and Polosukhin]{attention}
Ashish Vaswani, Noam Shazeer, Niki Parmar, Jakob Uszkoreit, Llion Jones, Aidan~N Gomez, {\L}ukasz Kaiser, and Illia Polosukhin.
\newblock Attention is all you need.
\newblock 2017.

\bibitem[Walch et~al.(2017)Walch, Hazirbas, Leal-Taixe, Sattler, Hilsenbeck, and Cremers]{lstmpose}
Florian Walch, Caner Hazirbas, Laura Leal-Taixe, Torsten Sattler, Sebastian Hilsenbeck, and Daniel Cremers.
\newblock Image-based localization using lstms for structured feature correlation.
\newblock In \emph{ICCV}, pages 627--637, 2017.

\bibitem[Wang et~al.(2019)Wang, Marcotte, and Olson]{lidar3}
Xipeng Wang, Ryan~J Marcotte, and Edwin Olson.
\newblock Glfp: Global localization from a floor plan.
\newblock In \emph{IROS}, pages 1627--1632, 2019.

\bibitem[Welch and Bishop(1995)]{kalmanFilter}
Greg Welch and Gary Bishop.
\newblock An introduction to the kalman filter.
\newblock Technical Report 95-041, University of North Carolina at Chapel Hill, 1995.

\bibitem[Wu et~al.(2017)Wu, Ma, and Hu]{branchpose}
Jian Wu, Liwei Ma, and Xiaolin Hu.
\newblock Delving deeper into convolutional neural networks for camera relocalization.
\newblock In \emph{ICRA}, pages 5644--5651, 2017.

\bibitem[Xia et~al.(2022)Xia, Booij, Manfredi, and Kooij]{crossview}
Zimin Xia, Olaf Booij, Marco Manfredi, and Julian~FP Kooij.
\newblock Visual cross-view metric localization with dense uncertainty estimates.
\newblock In \emph{ECCV}, pages 90--106, 2022.

\bibitem[Yao et~al.(2018)Yao, Luo, Li, Fang, and Quan]{mvsnet}
Yao Yao, Zixin Luo, Shiwei Li, Tian Fang, and Long Quan.
\newblock Mvsnet: Depth inference for unstructured multi-view stereo.
\newblock In \emph{ECCV}, pages 767--783, 2018.

\bibitem[Yao et~al.(2019)Yao, Luo, Li, Shen, Fang, and Quan]{rmvsnet}
Yao Yao, Zixin Luo, Shiwei Li, Tianwei Shen, Tian Fang, and Long Quan.
\newblock Recurrent mvsnet for high-resolution multi-view stereo depth inference.
\newblock In \emph{CVPR}, pages 5525--5534, 2019.

\bibitem[Zheng et~al.(2020)Zheng, Zhang, Li, Tang, Gao, and Zhou]{s3d}
Jia Zheng, Junfei Zhang, Jing Li, Rui Tang, Shenghua Gao, and Zihan Zhou.
\newblock Structured3d: A large photo-realistic dataset for structured 3d modeling.
\newblock In \emph{ECCV}, pages 519--535, 2020.

\bibitem[Zhou et~al.(2018)Zhou, Ummenhofer, and Brox]{deeptam}
Huizhong Zhou, Benjamin Ummenhofer, and Thomas Brox.
\newblock Deeptam: Deep tracking and mapping.
\newblock In \emph{ECCV}, pages 822--838, 2018.

\bibitem[Zhu et~al.(2021)Zhu, Yang, and Chen]{vigor}
Sijie Zhu, Taojiannan Yang, and Chen Chen.
\newblock Vigor: Cross-view image geo-localization beyond one-to-one retrieval.
\newblock In \emph{CVPR}, pages 3640--3649, 2021.

\end{thebibliography}
}


\end{document}